\documentclass{article}

\usepackage[utf8]{inputenc}

\usepackage{spconf}
\usepackage{amsmath,amssymb}
\usepackage{graphicx}

\usepackage{booktabs,multirow,array,makecell}
\usepackage{tabularx}
\newcolumntype{Y}{>{\centering\arraybackslash}X}

\usepackage[caption=false,font=footnotesize]{subfig}

\usepackage[table]{xcolor}
\definecolor{backblue}{RGB}{214,230,255}
\definecolor{backred}{RGB}{255,220,220}

\usepackage[hidelinks]{hyperref}
\usepackage[capitalise,noabbrev,nameinlink]{cleveref}

\newcommand{\blue}[1]{\textcolor{blue}{#1}}
\newcommand{\red}[1]{\textcolor{red}{#1}}

\title{R\textsuperscript{3}G: A Reasoning--Retrieval--Reranking Framework for Vision-Centric Answer Generation}

\name{%
\parbox{\linewidth}{\centering
Zhuohong Chen$^{1\dagger}$, Zhengxian Wu$^{1\dagger}$, Zirui Liao$^{1}$,
Shenao Jiang$^{1}$, Hangrui Xu$^{3}$, Yang Chen$^{1}$,\\
Chaokui Su$^{2}$, Xiaoyu Liu$^{2}$, Haoqian Wang$^{1}$%
\sthanks{$\dagger$ Equal contribution. * Corresponding author}
}}

\address{\fontsize{11}{11}\selectfont
$^{1}$The Shenzhen International Graduate School, Tsinghua University, China\\
$^{2}$State Key Laboratory of Nuclear Power Safety Technology and Equipment, China\\
$^{3}$School of Computer Science and Information Engineering, Hefei University of Technology, China
}

\begin{document}
\ninept
\maketitle

\begin{abstract}
Vision-centric retrieval for VQA requires retrieving images to supply missing visual cues and integrating them into the reasoning process. However, selecting the right images and integrating them effectively into the model’s reasoning remains challenging.
To address this challenge, we propose \textbf{R\textsuperscript{3}G}, a modular \textbf{Reasoning--Retrieval--Reranking} framework.
It first produces a brief reasoning plan that specifies the required visual cues, then adopts a two-stage strategy, with coarse retrieval followed by fine-grained reranking, to select evidence images.
On MRAG-Bench, R\textsuperscript{3}G improves accuracy across six MLLM backbones and nine sub-scenarios, achieving state-of-the-art overall performance. Ablations show that sufficiency-aware reranking and reasoning steps are complementary, helping the model both \emph{choose} the right images and \emph{use} them well. We release code and data at \url{https://github.com/czh24/R3G}.
\end{abstract}


\section{Introduction}
\label{sec:intro}


In recent years, multimodal retrieval-augmented-generation \cite{McKinzie2024MM1MA}\cite{Liu2023ImprovedBW} has demonstrated promising results in domains such as long-tail knowledge question answering\cite{Caffagni2024WikiLLaVAHR}\cite{Ling2025MMKBRAGAM}\cite{Tian2025CoReMMRAGCK} and document-based QA\cite{Cho2024M3DocRAGMR}\cite{Gong2025MMRAGDocQAAM}\cite{Hei2024DRRAGAD}.
By leveraging external knowledge bases as supplementary information, MLLMs are able to address questions that extend beyond their inherent knowledge boundaries\cite{Marino2019OKVQAAV}\cite{Schwenk2022AOKVQAAB}.
While these approaches have primarily utilized textual information as supplementary knowledge\cite{Mensink2023EncyclopedicVV}\cite{Chen2023CanPV}, recent research has begun to explore the use of visual information as an additional source for QA.

\begin{figure}[t]  
  \centering
  \includegraphics[width=\columnwidth]{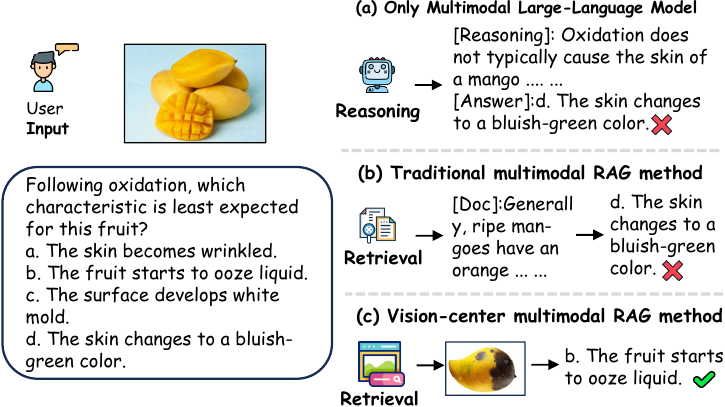} 
  \caption{Comparison of different methods for VQA.While MLLMs and text-based RAG often misalign with the question, vision-centric retrieval produces correct answers.}
  \vspace{-1.0em}
  \label{fig:fig1}
\end{figure}

Vision-Centric Retrieval for VQA needs to first retrieve additional images that provide missing visual cues and then integrate them for reasoning.
For example, in Fig.~\ref{fig:fig1} we ask: ``which characteristic is least expected for this fruit?''
"Answering this question is easier if the model can retrieve exemplars exhibiting the target state (e.g., images of rotten or browned mangoes) and then reason over the retrieved evidence.
Currently, the only available approach for this task is MRAG\cite{Hu2024MRAGBenchVE}, which ranks candidate images based on their global visual similarity to the query image.
However, such retrieval often fails in two ways. 
First, visually similar images can be semantically misaligned with the question intent.
Second, even when candidates are loosely on topic, their incidental content (backgrounds, lighting, unrelated objects) can dominate the model’s attention, distracting the answer generator from the correct visual cues\cite{Agrawal2017DontJA}\cite{Xiao2020NoiseOS}.

To address the limitations, we present \textbf{R\textsuperscript{3}G}, a modular \textbf{R}easoning– \textbf{R}etrieve–\textbf{R}erank framework for vision-centric VQA. 
R\textsuperscript{3}G first performs \emph{Reasoning-Before-Evidence} directly from the text–image query $(q_t,q_v)$ which can increase the model's attention to the correct visual cues, thereby improving the answer accuracy.
It then performs coarse retrieval from an external knowledge base to build a diverse top-$P$ pool.
This coarse stage serves to filter the large knowledge base into a small candidate pool, focusing on images that resemble the query in overall look or topic.
After that, an MLLM-as-Judge module, conditioned on $(q_t,q_v)$, evaluates each candidate.
The judge assigns three interpretable sub-scores (semantic relatedness, target correspondence, and answerability), which are aggregated into a fine-grained score. 
We then fuse this with the coarse retrieval score to rank candidates and select the top-$k$ evidence images that are visually similar to the query and semantically aligned with the question intent. 
We evaluate our approach on the only available vision-centric VQA benchmarks and observe consistent improvements across different MLLMs. 
On LLaVA-NeXT-Interleave-7B, our method achieves a 5.99\% gain in answer accuracy over mRAG, demonstrating its clear advantage. 
In addition, ablation studies confirm the effectiveness of both the Reasoning-Before-Evidence and the MLLM-as-Judge module, as well as their complementary effect.

We summarize our three contributions as follows:
\vspace{-1pt}
\begin{itemize}
  \item We propose R\textsuperscript{3}G, a Reasoning–Retrieval–Reranking framework that achieves new SOTA results on the vision-centric VQA dataset, improving downstream answer accuracy.

\begin{figure*}[t]  
\centering
\includegraphics[width=\textwidth]{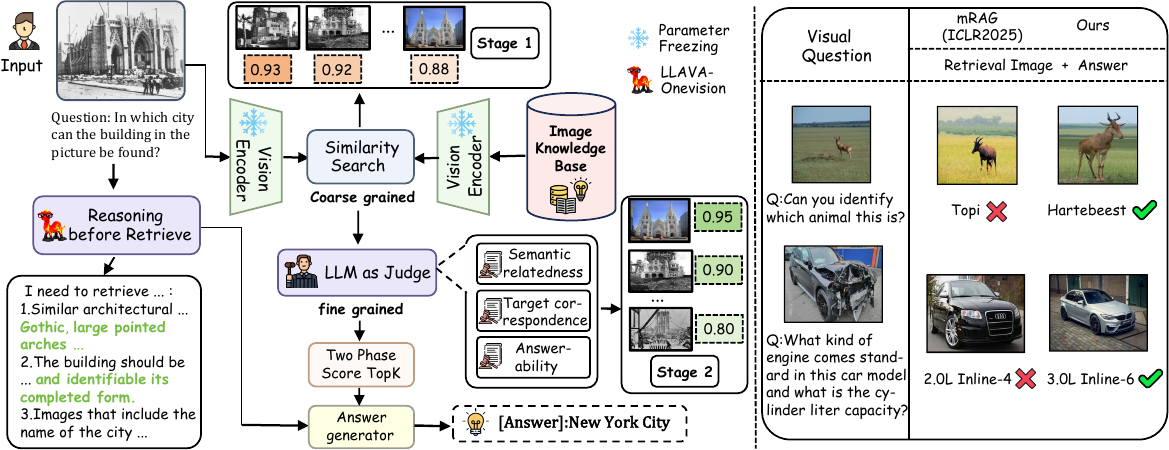}  
\caption{\textbf{Overview of R\textsuperscript{3}G for vision-centric VQA.}
\textbf{Left:}(i) A \emph{Reasoning-Before-Evidence} plan is produced from $(q_t,q_v)$ alone. 
(ii) Coarse image retrieval forms a candidate pool from an external image knowledge base. 
(iii) An \emph{MLLM-as-Judge} module reranks candidates with three criteria and fuses scores to select top-$k$ evidence images. 
(iv) The plan and selected images guide the MLLM to generate the final answer. 
\textbf{Right:} qualitative comparison with MRAG baseline.}
\vspace{-0.8em}
\label{fig:pipeline}
\end{figure*}

  \vspace{-1.5pt}
\item We design a two-stage image retrieval mechanism that fuses coarse retrieval with a fine grained, three criteria check based on semantic relatedness, target correspondence, and answerability to choose evidence images for answer generation.
  \vspace{-1.5pt}
  \item We study a \textit{Reasoning-Before-Evidence} strategy where the model, without retrieved images, states what is missing and which visual cues to check. This keeps later reasoning from being pushed off track by wrong or noisy images.
  \vspace{-0.125em}
\end{itemize}

\vspace{-0.9em}\section{Method}\vspace{-0.6em}
\noindent This section presents a modular \emph{reasoning-retrieval-reranking} framework for VQA. 
First, a \textit{Reasoning-Before-Evidence plan} (\S\ref{sec:cot}) is formed from \((q_t,q_v)\) alone, without using any retrieved images; this limits the influence of misleading evidence.
Next, the coarse stage (\S\ref{sec:retrieval}) computes Stage-1 scores for database images using the query image’s global semantics and ranks them to form a candidate pool. 
Then, an \textit{MLLM-as-Judge Reranking} module (\S\ref{sec:rerank}), conditioned on \((q_t,q_v)\), produces three sub-scores, aggregates them into a Stage-2 score, and fuses it with the Stage-1 score to balance coarse- and fine-grained signals. 
Finally, the top-\(k\) reranked images, together with the \textit{Reasoning-Before-Evidence} plan, are provided as supplementary input to the MLLM for answer generation.
This pipeline is concise, robust, and reproducible, and it improves both retrieval quality and downstream answer accuracy.

\vspace{-0.6em}\subsection{Reasoning-Before-Evidence}\vspace{-0.7em}
\label{sec:cot}
In multimodal RAG, we retrieve images as supplementary evidence for the MLLM, yet these images often contain noise. Processing such noise can bias the model’s reasoning and cause errors in downstream VQA. To mitigate this, we produce the reasoning \emph{before} using retrieved images, based only on $(q_t,q_v)$. Let $\mathcal{R}=\{r_i\}_{i=1}^{m}$ denote a list of single-sentence steps as follows:
{\setlength{\abovedisplayskip}{5pt}
 \setlength{\belowdisplayskip}{5pt}
 \setlength{\abovedisplayshortskip}{5pt}
 \setlength{\belowdisplayshortskip}{5pt}
\begin{equation}
\mathcal{R}^{\star}=\operatorname*{argmax}_{\mathcal{R}\in\mathbb{S}} \log p_\theta\!\big(\mathcal{R}\mid q_t,q_v\big),
\end{equation}}
where $\mathbb{S}$ is the set of step lists that are well-formed, aligned with the task, and verifiable (e.g., ordered visual checks). $\mathcal{R}^\star$ specifies what to inspect, in which order, and which cues require external evidence.

\vspace{-0.6em}\subsection{Coarse Image Retrieval}\vspace{-0.7em}
\label{sec:retrieval}
Given a text query $q_t$ and a query image $q_v$, we first apply the visual encoder $f_v:\mathcal{I}\!\to\!\mathbb{R}^{d}$. 
The encoder maps $q_v$ to an $\ell_2$-normalized feature $f_v(q_v)\in\mathbb{R}^d$. 
We then encode each image $I_i$ in the external image knowledge base $\mathcal{D}=\{I_i\}_{i=1}^{N}$ to obtain features $f_v(I_i)$. 
With these features, we compute cosine similarity (i.e., the inner product under $\ell_2$ normalization):
{\setlength{\abovedisplayskip}{5pt}
 \setlength{\belowdisplayskip}{5pt}
 \setlength{\abovedisplayshortskip}{5pt}
 \setlength{\belowdisplayshortskip}{5pt}
\begin{equation}
s_{\mathrm{img}}(q_v,I)=f_v(q_v)^\top f_v(I).
\end{equation}}
where $s_{\mathrm{img}}(q_v,I)$ is the similarity between $q_v$ and a candidate $I$, and $f_v(\cdot)$ denotes the visual embedding function.
We retain the top-$P$ candidates in descending order of $s_{\mathrm{img}}$:
{\setlength{\abovedisplayskip}{5pt}
 \setlength{\belowdisplayskip}{5pt}
 \setlength{\abovedisplayshortskip}{5pt}
 \setlength{\belowdisplayshortskip}{5pt}
\begin{equation}
\label{eq:topp}
\begin{aligned}
\mathbf{C}_P &= \big(I_{(1)},\ldots,I_{(P)}\big),\\
s_{\mathrm{img}}(q_v,I_{(1)}) &\ge \cdots \ge s_{\mathrm{img}}(q_v,I_{(P)}).
\end{aligned}
\end{equation}
}
We normalize the similarities within $\mathbf{C}_P$ using a temperature-scaled softmax to obtain probability-like weights:
{\setlength{\abovedisplayskip}{5pt}
 \setlength{\belowdisplayskip}{5pt}
 \setlength{\abovedisplayshortskip}{5pt}
 \setlength{\belowdisplayshortskip}{5pt}
\begin{equation}
\label{eq:stage1-def}
\begin{gathered}
\tilde{s}_{\mathrm{img}}(q_v,I)=
\frac{\exp\!\big(s_{\mathrm{img}}(q_v,I)/\tau\big)}
{\sum_{J\in\mathbf{C}_P}\exp\!\big(s_{\mathrm{img}}(q_v,J)/\tau\big)},\qquad \tau>0,\\
s_1(i) \triangleq \tilde{s}_{\mathrm{img}}(q_v, I_{(i)}).
\end{gathered}
\end{equation}
}
where the temperature $\tau$ is introduced to keep the first-stage scores within a controlled range.

Therefore, the coarse retrieval stage filters out semantically mismatched images and retains a subset from the large knowledge base. 
The selection operates on global visual semantics, resulting in a cleaner candidate pool that facilitates the fine-grained scoring and score fusion in~\S\ref{sec:rerank}.

\begin{table*}[t]
\caption{Accuracy on nine scenarios on MRAG-Bench( with EVA-CLIP retriever). The relative difference in performance compared to the score using \textbf{MRAG} is shown in the subscripts, with \blue{blue} indicating a decrease in performance and \red{red} indicating an increase.}
\label{tab:main-results}
\centering
 \renewcommand\tabcolsep{2.5pt} 
 \renewcommand\arraystretch{0.95} 
 \resizebox{1.0\linewidth}{!}{
\begin{tabular}{l@{}l|llll|llll|l|l}
\toprule
\multicolumn{2}{c|}{\multirow{2}{*}{Model}} &
\multicolumn{4}{c|}{Perspective} &
\multicolumn{4}{c|}{Transformative} &
\multicolumn{1}{c|}{\multirow{2}{*}{Others}} &
\multicolumn{1}{c}{\multirow{2}{*}{Overall}} \\
\cmidrule(lr){3-6} \cmidrule(lr){7-10}
& & \textbf{Angle} & \textbf{Partial} & \textbf{Scope} & \textbf{Occlusion}
& \textbf{Temporal} & \textbf{Deformation} & \textbf{Incomplete} & \textbf{Biological} & & \\
\midrule

\multirow{2}{*}{Mantis-8B-clip-llama3\cite{Jiang2024MANTISIM}} & +MRAG
& 36.65 & 34.96 & 42.16 & 47.22 & 50.34 & 33.33 & 18.63 & 21.57 & 42.50 & 36.88 \\
& +Ours
& 40.33$_\text{\textcolor{red}{+3.68}}$ & 45.13$_\text{\textcolor{red}{+10.17}}$ & 42.08$_\text{\textcolor{blue}{-0.08}}$ & 51.44$_\text{\textcolor{red}{+4.22}}$ & 51.42$_\text{\textcolor{red}{+1.08}}$ & 35.21$_\text{\textcolor{red}{+1.88}}$ & 19.03$_\text{\textcolor{red}{+0.40}}$ & 21.89$_\text{\textcolor{red}{+0.32}}$ & 50.58$_\text{\textcolor{red}{+8.08}}$ & 41.26$_\text{\textcolor{red}{+4.38}}$ \\
\midrule

\multirow{2}{*}{Mantis-8B-siglip-llama3\cite{Jiang2024MANTISIM}} & +MRAG
& 42.55 & 35.37 & 47.06 & 47.22 & 42.95 & 45.10 & 23.53 & 29.41 & 40.83 & 39.62 \\
& +Ours
& 48.41$_\text{\textcolor{red}{+5.86}}$ & 45.50$_\text{\textcolor{red}{+10.13}}$ & 46.18$_\text{\textcolor{blue}{-0.88}}$ & 50.08$_\text{\textcolor{red}{+2.86}}$ & 43.62$_\text{\textcolor{red}{+0.67}}$ & 47.32$_\text{\textcolor{red}{+2.22}}$ & 24.51$_\text{\textcolor{red}{+0.98}}$ & 29.99$_\text{\textcolor{red}{+0.58}}$ & 46.78$_\text{\textcolor{red}{+5.95}}$ & 42.36$_\text{\textcolor{red}{+2.74}}$ \\
\midrule

\multirow{2}{*}{Deepseek-VL-7B-chat\cite{Lu2024DeepSeekVLTR}} & +MRAG
& 33.54 & 32.11 & 33.33 & 37.04 & 43.62 & 40.20 & 20.59 & 26.47 & 45.00 & 34.66 \\
& +Ours
& 39.44$_\text{\textcolor{red}{+5.90}}$ & 45.12$_\text{\textcolor{red}{+13.01}}$ & 46.08$_\text{\textcolor{red}{+12.75}}$ & 50.93$_\text{\textcolor{red}{+13.89}}$ & 43.62$_\text{\textcolor{red}{+0.00}}$ & 33.33$_\text{\textcolor{blue}{-6.87}}$ & 24.51$_\text{\textcolor{red}{+3.92}}$ & 15.69$_\text{\textcolor{blue}{-10.78}}$ & 54.17$_\text{\textcolor{red}{+9.17}}$ & 40.28$_\text{\textcolor{red}{+5.62}}$ \\
\midrule

\multirow{2}{*}{LLaVA-NeXT-Interleave-7B\cite{Li2024LLaVANeXTInterleaveTM}} & +MRAG
& 40.06 & 33.33 & 39.22 & 56.48 & 43.62 & 44.12 & 27.45 & 36.27 & 49.17 & 40.35 \\
& +Ours
& 46.58$_\text{\textcolor{red}{+6.52}}$ & 48.37$_\text{\textcolor{red}{+15.04}}$ & 54.90$_\text{\textcolor{red}{+15.68}}$ & 61.11$_\text{\textcolor{red}{+4.63}}$ & 43.62$_\text{\textcolor{red}{+0.00}}$ & 46.08$_\text{\textcolor{red}{+1.96}}$ & 36.27$_\text{\textcolor{red}{+8.82}}$ & 25.49$_\text{\textcolor{blue}{-10.78}}$ & 50.83$_\text{\textcolor{red}{+1.66}}$ & 46.34$_\text{\textcolor{red}{+5.99}}$ \\
\midrule

\multirow{2}{*}{LLaVA-OneVision\cite{Li2024LLaVAOneVisionEV}} & +MRAG
& 50.93 & 48.78 & 50.00 & 60.19 & 50.34 & 48.04 & 33.33 & 53.92 & 54.17 & 50.11 \\
& +Ours
& 54.66$_\text{\textcolor{red}{+3.73}}$ & 60.98$_\text{\textcolor{red}{+12.20}}$ & 56.86$_\text{\textcolor{red}{+6.86}}$ & 62.96$_\text{\textcolor{red}{+2.77}}$ & 51.68$_\text{\textcolor{red}{+1.34}}$ & 52.94$_\text{\textcolor{red}{+4.90}}$ & 34.31$_\text{\textcolor{red}{+0.98}}$ & 54.90$_\text{\textcolor{red}{+0.98}}$ & 60.00$_\text{\textcolor{red}{+5.83}}$ & 55.14$_\text{\textcolor{red}{+5.03}}$ \\
\midrule

\multirow{2}{*}{Qwen2.5-VL-7B-Instruct\cite{Bai2025Qwen25VLTR}} & +MRAG
& 61.18 & 51.63 & 57.84 & 67.59 & 64.43 & 46.08 & 33.33 & 53.92 & 57.50 & 55.95 \\
& +Ours
& 67.70$_\text{\textcolor{red}{+6.52}}$ & 61.79$_\text{\textcolor{red}{+10.16}}$ & 64.71$_\text{\textcolor{red}{+6.87}}$ & 65.74$_\text{\textcolor{blue}{-1.85}}$ & 59.06$_\text{\textcolor{blue}{-5.37}}$ & 29.41$_\text{\textcolor{blue}{-16.67}}$ & 46.08$_\text{\textcolor{red}{+12.75}}$ & 51.96$_\text{\textcolor{blue}{-1.96}}$ & 56.67$_\text{\textcolor{blue}{-0.83}}$ & 58.61$_\text{\textcolor{red}{+2.66}}$ \\
\bottomrule
\end{tabular}
}
\vspace{-10pt}
\end{table*}

\vspace{-0.7em}\subsection{MLLM-as-Judge Reranking}\vspace{-0.7em}
\label{sec:rerank}

The coarse image retrieval stage selects candidates by the global visual semantics of the query image \(q_v\) but does not consider the text query \(q_t\) or verify answerability.
To remedy this limitation, we introduce an \emph{MLLM-as-Judge Reranking} module that is explicitly conditioned on \((q_t,q_v)\).
For each \(I_{(i)}\in\mathcal{C}_P\), the judge evaluates whether the candidate provides direct and reliable evidence for \((q_t,q_v)\) and then outputs a rationale--score pair:
{\setlength{\abovedisplayskip}{5pt}
 \setlength{\belowdisplayskip}{5pt}
 \setlength{\abovedisplayshortskip}{5pt}
 \setlength{\belowdisplayshortskip}{5pt}
\begin{equation}
  J_{\phi}:\ (q_t,q_v,I_{(i)}) \;\mapsto\; (\rho_i,\ s_2(i)),
  \label{eq:judge-map}
\end{equation}
}
where\(J_{\phi}\) is the {MLLM-as-Judge}, it returns a natural-language rationale \(\rho_i\) and an evidence-sufficiency score \(s_2(i)\).But a single score can be unstable across queries. To stabilize and structure the decision, the judge is required to output three observable sub-scores in [0,1], then aggregate them into the stage-2 score:
{\setlength{\abovedisplayskip}{5pt}
 \setlength{\belowdisplayskip}{5pt}
 \setlength{\abovedisplayshortskip}{5pt}
 \setlength{\belowdisplayshortskip}{5pt}
\begin{equation}
\label{eq:s2-from-subscores}
\begin{gathered}
s_2(i) = \lambda_r\, r_i + \lambda_t\, t_i + \lambda_a\, a_i,\\
\end{gathered}
\end{equation}
}
\noindent where \(\lambda_r,\lambda_t,\lambda_a\) are the weights of the three sub-scores, normalized to sum to unity \((\lambda_r+\lambda_t+\lambda_a=1)\); \(r_i\) represents semantic relevance, \(t_i\) represents target correspondence, and \(a_i\) represents answerability. They are generated by the module \textit{MLLM-as-Judge} according to \ specific scoring criteria as follows:
{\setlength{\abovedisplayskip}{5pt}
 \setlength{\belowdisplayskip}{5pt}
 \setlength{\abovedisplayshortskip}{5pt}
 \setlength{\belowdisplayshortskip}{5pt}
\begin{equation}
  (r_i,\, t_i,\, a_i) \;=\; J_{\phi}\big(q_t,\, q_v,\, I_{(i)};\rho_i,\, {G}\big)
\end{equation}
}

\noindent
where \({G}\) represents the scoring guidelines used to determine the three sub-scores.Specifically, they are as follows:

\noindent\textbf{Semantic relatedness.}\;
Semantic relatedness \(r_i\in[0,1]\) measure whether the dominant semantics of \(I_{(i)}\), such as category, part, scene, or action—match the intent of \((q_t,q_v)\). This score suppresses look-alike yet off-topic images in the candidate pool.

\noindent\textbf{Target correspondence.}\;
Target correspondence \(t_i\in[0,1]\) measure whether the evidence focuses on the exact target or scene asked by \(q_t\) with a comparable viewpoint, scale, and clarity, rather than providing only broad class context for the question.

\noindent\textbf{Answerability.}\;
Answerability \(a_i\in[0,1]\) measure whether combining \(I_{(i)}\) with the query image \(q_v\) makes the question decidable by supplying the missing cues (e.g., occluded configuration, temporal cue, fine texture) so that the correct answer exists.

To take into account both coarse-grained relevance and fine-grained sufficiency,  we balance the two stages' scores and obtain a ranking score for each candidate image.
{\setlength{\abovedisplayskip}{5pt}
 \setlength{\belowdisplayskip}{5pt}
 \setlength{\abovedisplayshortskip}{5pt}
 \setlength{\belowdisplayshortskip}{5pt}
\begin{equation}
  S(i)=s_1(i) +  s_2(i),
  \label{eq:fusion}
\end{equation}
}
The final evidence set ${I}^{\star}$ is obtained by selecting the top-$k$ candidates according to $S(i)$ from the candidate pool.

\noindent After the preceding steps yield \(\mathcal{R}^{\star}\) and \(I^{\star}\), the MLLM generates the answer \(y\) for the given query as follows:
{\setlength{\abovedisplayskip}{5pt}
 \setlength{\belowdisplayskip}{5pt}
 \setlength{\abovedisplayshortskip}{5pt}
 \setlength{\belowdisplayshortskip}{5pt}
\begin{equation}
y^{\star}=\operatorname*{argmax}_{y}\ \log p_\theta\!\big(y \mid q_t,q_v,I^\star,\mathcal{R}^{\star}\big).
\end{equation}
}
In this way, we can reduce the risk of MLLM being misled by wrong information and improve the accuracy of answers.

\section{Experiments}
\label{sec:experiments}
In this section, we introduce the dataset and task settings, and detail our experimental setup and metrics. We also demonstrate the gains achieved by our framework, R\textsuperscript{3}G, across various models and tasks. Finally, we present the results of our ablation experiments.

\subsection{Dataset}
We evaluate our R\textsuperscript{3}G framework on \textsc{MRAG-Bench}\cite{Hu2024MRAGBenchVE}, which is currently the only benchmark targeting vision-centric VQA.
The benchmark contains nine sub-scenarios grouped into three families: 
\emph{Perspective} (Angle, Partial, Scope, Occlusion), 
\emph{Transformative} (Temporal, Deformation, Incomplete, Biological), 
and \emph{Others}. 
These scenarios explicitly stress cases where the query image lacks decisive information and external visual evidence is required.

\subsection{Implementation Details}
\textbf{Backbones.} We test six open-source MLLMs: Mantis-8B-clip-llama3, Mantis-8B-siglip-llama3\cite{Jiang2024MANTISIM}, DeepSeek-VL-7B-chat\cite{Lu2024DeepSeekVLTR}, LLaVA-NeXT-Interleave-7B\cite{Li2024LLaVANeXTInterleaveTM}, LLaVA-OneVision\cite{Li2024LLaVAOneVisionEV}, and Qwen\\-2.5VL-7B-Instruct\cite{Bai2025Qwen25VLTR}. 
All backbones are frozen and share the same prompt, image preprocessing, and decoding limits.\\
\textbf{Settings.} We report three settings that are comparable across backbones: 
(i) with MRAG (using the image retrieved by the MRAG baseline retriever); 
(ii) with Ours (R\textsuperscript{3}G framework, two-stage retrieval with fine-grained reranking and reasoning-chain guided generation as in Section~2). 
External evidence is injected as images.
The Judge’s aggregation weights in Eq.~(6) are fixed to $\lambda_{r}{=}0.20$, $\lambda_{t}{=}0.35$, and $\lambda_{a}{=}0.45$.
Due to paper space constraints, the full prompts for \emph{Reasoning-Before-Evidence} and \emph{MLLM-as-Judge Reranking} are provided in our code repository.\\
\textbf{Metrics.} Downstream performance is measured by \emph{Accuracy}. Retrieval is evaluated by \emph{Recall@K}.($K{=}1,3,5$), which checks whether the top–$K$ set contains at least one ground–truth evidence image.

\subsection{Main Results}
Across all backbones and sub-scenarios, R\textsuperscript{3}G consistently outperforms the MRAG baseline (Table~\ref{tab:main-results}). We report absolute accuracies with deltas relative to MRAG in the subscripts.

\textbf{Backbone-level analysis.}
For every backbone, adopting R\textsuperscript{3}G yields higher Overall accuracy than MRAG, with gains of roughly +2.7 to +6.0 percentage points. The trend holds for CLIP- and SigLIP-based Mantis models, as well as for LLaVA, DeepSeek-VL, and Qwen, indicating that the Reasoning-Retrieval–Reranking pipeline is broadly applicable across architectures. While most sub-scenarios improve, a few categories (e.g., \emph{Biological} or \emph{Temporal} for some models) show small regressions, suggesting room to further refine evidence selection in future iterations of R\textsuperscript{3}G.

\textbf{Scenario-level analysis.}
In \emph{Perspective}, improvements concentrate on \emph{Angle}, \emph{Partial}, and \emph{Occlusion}, where reranking filters visually similar but irrelevant candidates and prioritizes images that reveal the missing cues. 
In \emph{Transformative}, we keep only candidates with supporting evidence. This reduces noise and improves accuracy, especially on \emph{Deformation} and \emph{Incomplete}, though \emph{Biological} remains difficult.
For \emph{Others}, gains are modest, suggesting that purely visual evidence saturates quickly and may benefit from lightweight textual knowledge in future work.

\begin{table}[t]
\caption{Effect of Stage~1 and Stage~2 retrieved image counts on accuracy. Here, $p$ is Stage~1 coarse pool size and $k$ is Stage~2 reranked images injected; results are reported per scenario and overall.}
\label{tab:pk_retrieval}
\centering
\begin{minipage}{0.5\textwidth}
\centering
{\scriptsize
\setlength{\tabcolsep}{3pt}
\renewcommand{\arraystretch}{1.05}
\begin{tabular}{cc|cc|cc|c|c}
\toprule
\multicolumn{1}{c}{\multirow{2}{*}{\textbf{p}}} &
\multicolumn{1}{c|}{\multirow{2}{*}{\textbf{k}}} &
\multicolumn{2}{c|}{\textbf{Perspective}} &
\multicolumn{2}{c|}{\textbf{Transformative}} &
\multicolumn{1}{c|}{\multirow{2}{*}{\textbf{Others}}} &
\multicolumn{1}{c}{\multirow{2}{*}{\textbf{Overall}}} \\
\cmidrule(lr){3-4} \cmidrule(lr){5-6}
& & \textbf{Angle} & \textbf{Partial} & \textbf{Incomplete} & \textbf{Biological} & & \\
\midrule
1 & 1 & 47.26 & 52.27 & 29.59 & 46.91 & 51.77 & 47.43 \\
3 & 1 & 53.80 & 60.50 & 33.65 & 54.30 & 59.44 & 54.43 \\
3 & 3 & 53.30 & 58.91 & 33.33 & 53.05 & 58.26 & 53.45 \\
5 & 1 & 54.66 & 60.98 & 34.31 & 54.90 & 60.00 & 55.14 \\
5 & 3 & 53.92 & 60.58 & 33.86 & 54.46 & 59.46 & 54.59 \\
5 & 5 & 52.36 & 58.09 & 32.91 & 52.22 & 57.35 & 52.66 \\
\bottomrule
\end{tabular}
}
\end{minipage}
\vspace{-15pt}
\end{table}

\begin{table}[t]
\caption{\textbf{Ablations of the retrieval module.}
(a) Retrieval quality measured by \emph{Recall@K}: comparison between mRAG and ours.
(b) Effect of different image retrievers on downstream answer \emph{Accuracy}.}
\label{tab:retrieval_ablation_pair}

{\scriptsize
\setlength{\tabcolsep}{0pt}
\renewcommand{\arraystretch}{1.0}

\noindent
\begin{minipage}[t]{0.48\linewidth}
\raggedright
\setlength{\tabcolsep}{2.5pt}
\renewcommand{\arraystretch}{1.05}

\begin{tabular}{@{}lcc@{}}
\toprule
 & \textbf{Method} & \textbf{Recall@K} \\
\midrule
\multirow{2}{*}{K=1} & MRAG & 33.12 \\
                     & Ours & 37.86 \\
\midrule
\multirow{2}{*}{K=3} & MRAG & 58.63 \\
                     & Ours & 64.97 \\
\midrule
\multirow{2}{*}{K=5} & MRAG & 72.41 \\
                     & Ours & 77.15 \\
\bottomrule
\end{tabular}

\vspace{2pt}
{\footnotesize\itshape (a) MRAG vs Ours (R@K)}
\end{minipage}
\hfill
\begin{minipage}[t]{0.48\linewidth}
\setlength{\tabcolsep}{1.8pt}
\renewcommand{\arraystretch}{1.10}

\newsavebox{\tabB}
\sbox{\tabB}{%
  \begin{tabular}{@{}c|cccc@{}}
  \toprule
  \textbf{Scenario} & \textbf{EVA-CLIP} & \textbf{BLIP} & \textbf{CLIP} & \textbf{UniIR} \\
  \midrule
  \textbf{Partial}        & 60.98 & 58.98 & 58.79 & 58.52 \\
  \textbf{Scope}          & 56.86 & 55.03 & 54.83 & 54.51 \\
  \textbf{Deformation}    & 52.94 & 51.06 & 50.82 & 50.53 \\
  \textbf{Perspective}    & 58.10 & 56.82 & 56.51 & 56.21 \\
  \textbf{Transformative} & 48.79 & 48.23 & 48.02 & 47.71 \\
  \textbf{Others}         & 60.00 & 58.49 & 58.31 & 58.01 \\
  \textbf{Overall}        & 55.14 & 54.13 & 53.88 & 53.56 \\
  \bottomrule
  \end{tabular}
}

\makebox[\linewidth][r]{\usebox{\tabB}}

\vspace{2pt}
\makebox[\linewidth][r]{\makebox[\wd\tabB][c]{\footnotesize\itshape (b) Different image retrievers}}
\end{minipage}

} 
\end{table}

\begin{table}[t]
\caption{Ablation of module combinations. Fine Reranking(\emph{w.o.\ G}): judge relies only on MLLM's parametric knowledge; Fine Reranking(\emph{w.\ G}):judge follows explicit Guidelines. \emph{R*}: answer with question-conditioned reasoning steps.}
\label{tab:ablation_simple}
\centering
\begin{minipage}{0.45\textwidth}
\centering
{\scriptsize
\setlength{\tabcolsep}{3pt}
\renewcommand{\arraystretch}{1.10}
\begin{tabular}{ccc|ccc}
\toprule
\multicolumn{3}{c|}{\textbf{Module}} & \multicolumn{3}{c}{\textbf{Accuracy}} \\
\cmidrule(lr){1-3}\cmidrule(lr){4-6}
\makecell[c]{\textbf{Fine Reranking}\\\textbf{w.o.\ G}} &
\makecell[c]{\textbf{Fine Reranking}\\\textbf{w.\ G}} &
\textbf{R*} &
\textbf{Perspective} & \textbf{Transformative} & \textbf{Overall} \\
\midrule
  -& - & - & 54.63 & 38.24 & 48.48 \\
$\checkmark$ & - & - & 55.78 & 40.01 & 49.67 \\
 --& $\checkmark$ & - & 57.20 & 46.59 & 53.81 \\
 -- & -- & $\checkmark$ & 57.46 & 47.13 & 54.12 \\
-- & $\checkmark$ & $\checkmark$ & 58.10 & 48.79 & 55.14 \\
\bottomrule
\end{tabular}
}
\end{minipage}
\vspace{-6pt}
\end{table}

\vspace{-5pt}\subsection{Ablation Studies}

All ablations are conducted on MRAG- Bench.
Except for Table~\ref{tab:retrieval_ablation_pair}(b), which changes the first- stage retriever, we use \textbf{EVA- CLIP} for retrieval in all experiments.
We study four aspects:

\textbf{Evidence- count sensitivity (Table~\ref{tab:pk_retrieval}).}
We vary the Stage~1 pool $p$ and the Stage~2 verified set $k$.
Enlarging $p$ consistently improves accuracy: with $k{=}1$, moving from $p{=}1$ to $p{=}3$ yields a clear gain, and $p{=}5$ attains the best Overall (55.14\%).
By contrast, increasing $k$ at fixed $p$ provides little benefit and can reduce accuracy, indicating that a small verified set is preferable.
Overall, a \emph{wide} Stage~1 and a \emph{compact} Stage~2 work best; we adopt $p{=}5,\,k{=}1$ as a practical default, which is especially helpful for the \emph{Transformative} scenarios.

\textbf{Top-K Evidence Recall (R@K; Table~\ref{tab:retrieval_ablation_pair}a).}
We measure retrieval coverage by Recall@K, counting a query as recalled if any ground-truth evidence appears in the top-$K$. 
Because the pipeline is from coarse to fine, we fix the Stage-1 pool at $p=5$. 
Across $K\in\{1,3,5\}$, our method consistently exceeds the MRAG baseline, with the largest gain at $K{=}3$. 
This shows that reranking concentrates useful evidence near the top; given the MLLM’s limited image budget, prioritizing early placement is more effective than enlarging the Stage~2 input set (cf. Table~\ref{tab:pk_retrieval}).

\textbf{Different image retrievers (Table~\ref{tab:retrieval_ablation_pair}b).}
Changing the Stage- 1 retriever while keeping all other components fixed leads to modest changes.
EVA- CLIP\cite{Sun2023EVACLIPIT} attains the best Overall accuracy (55.14\%), followed by BLIP \cite{Li2022BLIPBL}(54.13\%), CLIP (53.88\%), and UNIir\cite{Wei2023UniIRTA} (53.56\%).
Per- scenario trends are consistent: EVA- CLIP is slightly better on \emph{Perspective} (58.10\%) and \emph{Transformative} (48.79\%), but the gaps are small across retrievers.
Since retrieval is used only for coarse selection and is followed by verification, the downstream performance is dominated by how evidence is checked and used, rather than by small differences among same- modality retrievers.

\textbf{Module-level ablations (Table~\ref{tab:ablation_simple}).}
Both fine reranking variants (\emph{with} and \emph{without} guidelines) outperform the base system, and the guideline-based variant delivers the larger gain (Overall: 49.67\%\,$\mathrm{vs.}$\,53.81\%).
Adding the question-conditioned reasoning module \(\mathcal{R}^{\star}\) yields a further improvement, and combining guideline-based reranking with \(\mathcal{R}^{\star}\) attains the best Overall (55.14\%).
These results suggest complementarity: reranking strengthens evidence selection, whereas \(\mathcal{R}^{\star}\) improves the use of that evidence during answering.
To isolate the effect of score fusion, we compare three variants under the same setup: (i) \emph{Stage-1 only} (54.12\% Overall), (ii) \emph{Stage-2 only} (selecting inputs solely by Stage-2 scores without fusion) achieving 57.97\% on \emph{Perspective}, 48.57\% on \emph{Transformative}, and 54.99\% Overall (numbers not listed in Table~\ref{tab:ablation_simple}), and (iii) the \emph{fused} model, which reaches 55.14\% Overall.
The results show that the fine-grained information emphasized in Phase 2 produced a larger improvement in answer accuracy than the global semantic information emphasized in Phase 1, and fusing coarse and fine information is more reliable than than using either stage alone.

\section{Conclusion}
\label{sec:conclusion}
In this paper, we present \textbf{R\textsuperscript{3}G}, a reasoning- retrieval- rerank framework for vision- centric VQA.
It targets cases where the query image lacks key visual cues.
R\textsuperscript{3}G has two main contributions.
First,  to prevent noisy retrieved images from steering the model’s reasoning, we generate a question- conditioned chain of thought before providing the retrieved images to the model.
This plan then guides the MLLM’s reasoning and answer generation.
Second, it adopts a two- retrieval and scoring strategy: a coarse retrieval stage that filters highly dissimilar images and reduces the search space to a small candidate set, followed by a fine grained reranking stage that scores candidates by question- conditioned relevance.
On MRAG-Bench, R\textsuperscript{3}G improves performance across six MLLM backbones and nine scenarios, achieving SOTA overall accuracy.
These results suggest that effective VQA requires both \emph{choosing} the right images and \emph{using} them well, and R\textsuperscript{3}G serve as a useful reference framework. 

\section{ACKNOWLEDGEMENTS}
This work is supported by the NSFC fund (62576190), in part by the Shenzhen Science and Technology Project under Grant (KJZD20240903103210014, JCYJ20220818101001004)


\bibliographystyle{IEEEbib}
\bibliography{refs}

\end{document}